%% file: icme2021template.tex
\documentclass{article}

\usepackage{amsmath,epsfig}
\usepackage[preprint]{spconfa4}

\copyrightnotice{978-1-6654-3864-3/21/\$31.00 ©2021 IEEE}

\let\OLDthebibliography\thebibliography
\renewcommand\thebibliography[1]{
  \OLDthebibliography{#1}
  \setlength{\parskip}{0pt}
  \setlength{\itemsep}{0pt plus 0.3ex}
}

\usepackage{xcolor}
\usepackage{verbatim}  
\usepackage{amssymb}
\usepackage{multirow}
\usepackage{bbm} 
\usepackage{enumitem}
\usepackage{hyperref}
\usepackage{url}

\begin{document}\sloppy

\def\x{{\mathbf x}}
\def\L{{\cal L}}

\title{Weakly-Supervised Image Semantic Segmentation Using Graph Convolutional Networks}
%
\name{Shun-Yi Pan$^{\ast}$\thanks{$^{\ast}$Both authors contributed equally to the paper}\thanks{ This work is supported by National Center for High-performance Computing, Taiwan.}, Cheng-You Lu$^{\ast}$, Shih-Po Lee, Wen-Hsiao Peng}
\address{National Yang Ming Chiao Tung University, Taiwan\\
\{sypan.cs05g, johnny305.cs04, mapl0756051.cs07g\}@nctu.edu.tw\\
wpeng@cs.nctu.edu.tw}

\maketitle

\pagenumbering{gobble}

\input{abs}

\vspace{-.5em}
\input{intro}

\vspace{-.5em}
\input{related}
\input{method}
\input{experiment}
\input{conclusion}

\bibliographystyle{IEEEbib}
\bibliography{reference}
\end{document}

%% file: abs.tex
\begin{abstract}
\vspace{-.25em}
This work addresses weakly-supervised image semantic segmentation based on image-level class labels. One common approach to this task is to propagate the activation scores of Class Activation Maps (CAMs) using a random-walk mechanism in order to arrive at complete pseudo labels for training a semantic segmentation network in a fully-supervised manner. However, the feed-forward nature of the random walk imposes no regularization on the quality of the resulting complete pseudo labels. To overcome this issue, we propose a Graph Convolutional Network (GCN)-based feature propagation framework. We formulate the generation of complete pseudo labels as a semi-supervised learning task and learn a 2-layer GCN separately for every training image by back-propagating a Laplacian and an entropy regularization loss. Experimental results on the PASCAL VOC 2012 dataset confirm the superiority of our scheme to several state-of-the-art baselines. \textcolor{black}{Our code is available at \url{https://github.com/Xavier-Pan/WSGCN}.}
 \end{abstract}

%% file: intro.tex
\section{Introduction}\label{sec:intro}
\vspace{-.5em}

Image semantic segmentation aims for classifying pixels in an image into their semantic classes. Training a semantic segmentation network often requires costly pixel-wise class labels. To avoid the time-consuming annotation process, weakly-supervised learning is introduced to utilize relatively low-cost weak labels such as bounding boxes, scribbles, and image-level class labels~\cite{MDC18,DSRG18, PSA18,IRNet19,SEAM20,SCE20,SingleStage20}. 


Recent research has been focused on weakly-supervised image semantic segmentation based on image-level class labels. The work in~\cite{PSA18} is representative of the popular two-stage framework. It begins with producing pseudo labels, followed by training a semantic segmentation network supervisedly using these labels. The generation of pseudo labels usually proceeds in three steps. The first step predicts crude estimates of labels, known as \textit{partial pseudo labels}, by using Class Activation Maps (CAMs)~\cite{CAM16}. They are partial in that only some pixels will receive their pseudo labels. In the second step, these partial pseudo labels are utilized to train an affinity network, requiring that pixels sharing the same label should have similar features. Lastly, the affinity network is applied to evaluate a Markov transition matrix for propagating the activation scores of CAMs through a random-walk mechanism, with the aim of producing \textit{complete pseudo labels} for all the pixels. On the basis of this two-stage framework, some works~\cite{SEAM20,SCE20} improve on initial CAMs while others~\cite{PSA18,IRNet19} attempt to learn better affinity matrices.

This paper addresses the propagation mechanism through learning a graph neural network.
To derive complete pseudo labels, all the prior works~\cite{PSA18,IRNet19,SEAM20,SCE20} rely on the random-walk mechanism to propagate the activation scores of CAMs as a form of label propagation. Due to its feed-forward nature, there is no regularization imposed on the resulting complete pseudo labels, the quality of which depends highly on the Markov matrix. Departing from label propagation, we propose a feature propagation framework based on learning a Graph Convolutional Network (GCN)~\cite{GCN17} for every training image. Our contributions include: 

 \begin{enumerate}[topsep=0pt]
  \setlength{\itemsep}{0pt}
  \setlength{\parskip}{0pt}
 \item We propagate the high-level semantic features of image pixels on a graph using a 2-layer graph convolutional network, followed by decoding the propagated features into semantic predictions.
 \item We cast the problem of obtaining complete pseudo labels from partial pseudo labels as a semi-supervised learning problem.
 \item We learn a separate GCN for every training image by back-propagating a Laplacian loss and an entropy loss to ensure the consistency of semantic predictions with image spatial details.
\end{enumerate}

Experimental results show that our complete pseudo labels have higher accuracy in Mean Intersection over Union (mIoU) than label propagation~\cite{PSA18,IRNet19}. The net effect is that the semantic segmentation network trained with our complete pseudo labels outperforms the state-of-the-art baselines~\cite{SEAM20,SCE20,SingleStage20,RRM20,MCIS20,WSSSGNN21} on the  PASCAL VOC 2012  dataset. 


%% file: related.tex
\begin{figure*}[t]
    \centering\footnotesize
    \includegraphics[width=.9\textwidth]{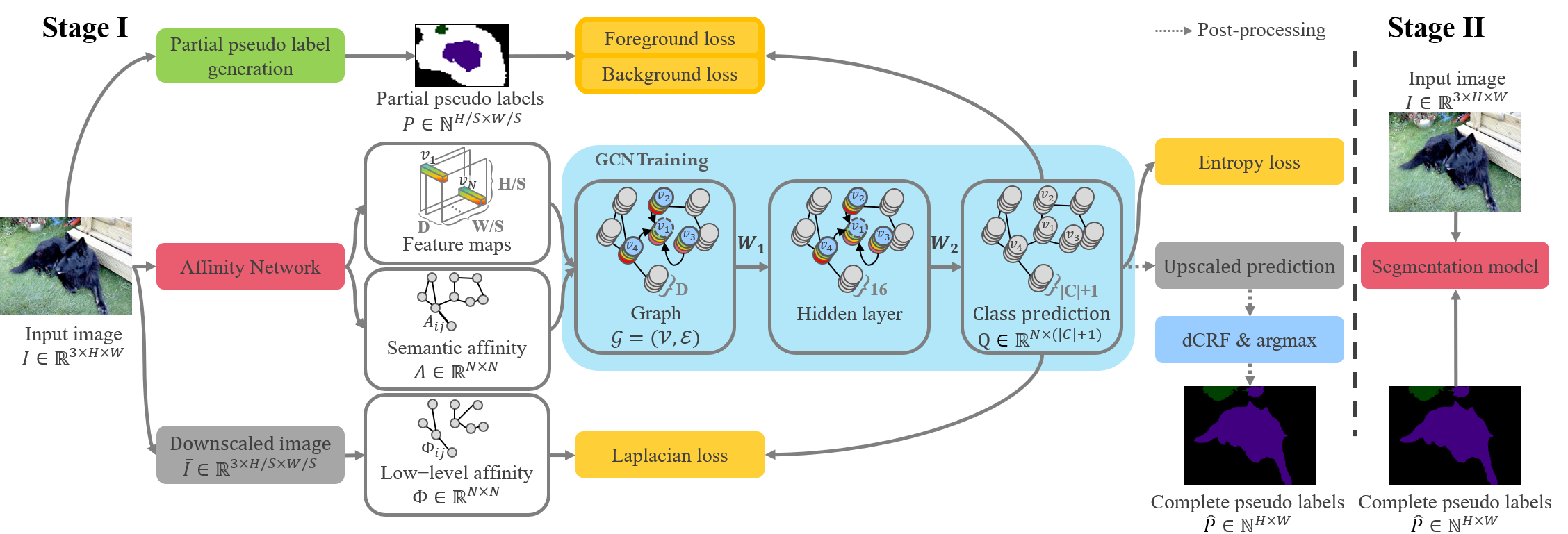}
    \caption{Overview of the proposed 2-stage weakly-supervised image semantic segmentation framework with GCN-based feature propagation. In our implementation, the Affinity Network is from \cite{PSA18,IRNet19}. 
    }
    \label{fig:overview}
\end{figure*}
\section{Related Work}\label{sec:Work}
\vspace{-.6em}

This section surveys prior works that use image-level class labels for weakly-supervised semantic segmentation, with a particular focus on the improvement of CAMs and how pseudo labels are derived from CAMs. 

\subsection{Class Activation Maps}\label{sec:PPLI}



The CAMs~\cite{CAM16} derived from image-level class labels often serve as initial seeds for generating partial pseudo labels. Many efforts have been devoted to improving the quality of CAMs. RRM~\cite{RRM20} learns jointly an image classifier and a semantic segmentation network, with the hope of regularizing the feature extraction towards better CAM generation. By the same token, SEAM~\cite{SEAM20} imposes an equivariant constraint on learning the feature extractor, requiring that the CAMs of two input images related through an affine transformation follow the same affine relationship. SC-CAM~\cite{SCE20} divides the object category into sub-categories and learns a feature extractor that can identify find-grained parts of an object. \textcolor{black}{GroupWSSS~\cite{WSSSGNN21} learns a graph neural network to leverage the semantic relations among images which share class labels partially for better CAM generation.}


\subsection{Generation of Complete Pseudo Labels}\label{sec:PL}
There are approaches~\cite{DSRG18,PSA18,IRNet19} that focus on generating complete pseudo labels from CAMs. DSRG~\cite{DSRG18} expands the partial pseudo labels by annotating iteratively the unlabeled pixels adjacent to the labeled ones through a semantic segmentation network. The propagation process, however, does not take into account the affinity between pixels. By contrast, PSA~\cite{PSA18} learns an affinity network to guide the propagation of the CAM scores by weighting differently the edges connecting adjacent pixels. The affinity network is trained by minimizing the $\l_1$ norm between the feature vectors of neighboring pixels that share the same semantic class according to their partial pseudo labels. \textcolor{
black}{IRNet~\cite{IRNet19} extends the idea to incorporate the boundary map in determining the affinity between pixels. Specifically, pixels separated by a strong boundary are considered semantically dissimilar, and vice versa.} SEAM~\cite{SEAM20} and SC-CAM~\cite{SCE20} adopt the same propagation framework for CAMs as~\cite{PSA18, IRNet19} to produce complete pseudo labels.
 
In common, most of these methods build the label propagation on a random-walk mechanism. The feed-forward nature of the random walk does not provide any guarantee on the quality of the complete pseudo labels \textcolor{black}{and does not consider low-level features during propagation.} Our work presents the first attempt at replacing the random walk with a GCN-based feature propagation scheme. 

%% file: method.tex
\section{Proposed Method} \label{section:method}
This work introduces a GCN-based feature propagation scheme, with the aim of predicting the semantic label for every pixel to produce complete pseudo labels. 
Unlike the random walk, which relies solely on the affinity between pixels in the feature domain to propagate the activation scores of CAMs, our scheme learns a GCN by regularizing the feature propagation with not only the aforementioned affinity information but also the color information of the input image. Furthermore, recognizing that the generation of pseudo labels is an off-line process, we train a separate GCN to optimize feature propagation for every training image. \textcolor{black}{We choose GCN instead of the convolutional neural networks because of the irregular affinity relations among feature samples.} 
 
\subsection{System Overview}
 Fig.~\ref{fig:overview} illustrates the overall architecture of the proposed method. As shown, the process proceeds in two stages: (1) the generation of pseudo labels, and (2) the training of a semantic segmentation network. \textcolor{black}{In the first stage, we follow \cite{PSA18, RRM20} to generate partial pseudo labels $P\in \mathbb{N}^{H/S \times W/S}$ for an input image $I\in \mathbb{R}^{3\times H\times W}$ of width $W$ and height $H$. Note that $P$ is the same size as the CAMs with $S$ denoting the downscaling factor. At the location $(x,y)$, the pseudo label $P(x,y)$ is assigned either a class label $c \in C \cup \{c^{bg}\}$ or $ignored$, where $C$ is the set of foreground classes, $c^{bg}$ denotes the background class, and the $ignored$ signals the unlabeled state.} Given the partial pseudo labels $P$, we consider the generation of complete pseudo labels $\hat{P} \in \mathbb{N}^{H \times W}$ to be a semi-supervised learning problem on a graph. The output of the first stage then comprises the complete pseudo labels for the image $I$, which are utilized in the second stage as ground-truth labels for training the semantic segmentation network. The following details the operation of each component.
 






\subsection{Inference of Complete Pseudo Labels on a Graph}
We begin by defining the graph structure. Let $\mathcal{G}=(\mathcal{V},\mathcal{E})$ be a graph, where $\mathcal{V}=\{v_1,...,v_{N} \}$ is a collection of nodes and $\mathcal{E}=\{A_{i,j}| v_i, v_j \in \mathcal{V}\}$ specify edges connecting nodes $v_i$ and $v_j$ with weights $A_{i,j}$. In our task, a node $v_i \in \mathbb{R}^{D}$ in $\mathcal{V}$ refers collectively to the co-located feature samples at $(x_i,y_i)$ in $D$ feature maps, each being of size $H/S \times W/S$ with the total number of nodes $N = H/S  \times W/S$. The choice of node features is detailed in Section~\ref{sec:setup}. The pseudo label of $v_i$ is denoted as $p_i = P(x_i,y_i)$. The edge weight $A_{i,j}$ measures the affinity between $v_i$ and $v_j$. Because GCN is amenable to a wide choice of affinity measures, we test two different measures \cite{PSA18,IRNet19} in our experiments (Section~\ref{sec:setup}). 

To generate complete pseudo labels $\hat{P}$, we perform feature propagation and inference on the graph $\mathcal{G}$ through a 2-layer GCN. The feed-forward inference proceeds as follows: 
\begin{equation}\label{eq:GCN}
Q = \sigma_s(\Tilde{A}(\sigma_r(\Tilde{A} V W_1)W_2),
\end{equation} where $V=[v_1, v_2, ... ,v_{N}]^T \in \mathbb{R}^{N \times D}$ is a matrix of node feature vectors, each of which is  $D$-dimensional; $W_1\in \mathbb{R}^{D \times 16},W_2 \in \mathbb{R}^{16 \times (|C|+1)}$ are the two learnable network parameters; $\sigma_r(\cdot),\sigma_s(\cdot)$ correspond to the ReLU and the softmax activation functions, respectively; and $\Tilde{A}=A+I_N$ is the sum of the affinity matrix $A \in \mathbb{R}^{N \times N}$ and an identity matrix $I_N$ of size $N$.
\textcolor{black}{Note that the number of classes which include background class is denoted as $|C|+1$.}
In the resulting matrix $Q=[q_1, q_2, ... ,q_{N}]^T \in \mathbb{R}^{N \times (|C|+1)}$, each row $q_i^T,i=1,2,\ldots,N$ signals the probability distribution of semantic classes \textcolor{black}{at pixel $(x_i,y_i)$ in the feature domain}. These probability distributions are then interpolated spatial-wise (using bilinear interpolation) to arrive at a full-resolution semantic prediction map, followed by applying the dCRF~\cite{CRF} in a channel-wise manner and taking the maximum across channels at every pixel for complete pseudo labels $\hat{P}$.

\subsection{Training a GCN for Feature Propagation}\label{sec_loss}
The training of \textcolor{black}{a GCN} for every image is formulated as a semi-supervised learning problem and incorporates four loss functions, namely the (1) foreground loss $\ell_{fg}$, (2) background loss $\ell_{bg}$, (3) entropy loss $\ell_{ent}$, and (4) Laplacian loss $\ell_{lp}$:
\begin{equation}
\label{eq:loss_all}
 \ell =  \ell_{fg}+\ell_{bg}+\beta_1 \ell_{ent}+\beta_2 \ell_{lp},
\end{equation}
where $\beta_1$ and $\beta_2$ are hyper-parameters. 
The first two are evaluated as the sum of cross entropies over the foreground and background pixels in the feature domain, respectively, where the foreground pixels have their partial pseudo labels $P(x,y) \in C$ while the background pixels have $P(x,y)=c^{bg}$. The rationale behind the separation of the cross entropies into the foreground and background groups is to address the imbalance between these two classes of pixels. In symbols, we have

\begin{equation}
\label{eq:loss_fg}
\ell_{fg} = -\frac{1}{|\mathcal{V}_{fg}|} \sum_{i \in \mathcal{V}_{fg}}{ \log{(q_i)_{p_i}} },
\end{equation}
\begin{equation}
\label{eq:loss_bg}
\ell_{bg} = -\frac{1}{|\mathcal{V}_{bg}|} \sum_{i \in \mathcal{V}_{bg}}{ \log{(q_i)_{p_i}} },
\end{equation}
\textcolor{black}{where $q_i$ from the $Q$ in Eq.~(\ref{eq:GCN}) denotes the predicted class distribution at pixel $i$; $(q_i)_{p_i}$ refers specifically to the predicted probability of the class corresponding to the partial pseudo label $p_i$; $\mathcal{V}_{fg}$ and $\mathcal{V}_{bg}$ are the foreground and background pixels according to the partial pseudo labels $P$.}

 For those pixels in the feature domain with their pseudo labels $P(x,y)$ marked as $\textit{ignored}$, i.e., unlabeled pixels, we impose the following entropy loss, requiring that the uncertainty about their class predictions should be minimized. In other words, it encourages the class predictions $q_i$ at those unlabeled pixels to approximate one-hot vectors. 
 
\begin{equation}
\label{eq:loss_ent}
 \ell_{ent} = -\frac{1}{|\mathcal{V}_{ig}|} \sum_{i \in \mathcal{V}_{ig}}\sum_{c\in \bar{C}}{(q_i)_{c}\log{(q_i)_{c}}},
\end{equation}
where $\bar{C}=C\cup \{c^{bg}\}$ and $\mathcal{V}_{ig}$ refers to unlabeled pixels.
In addition, motivated by the observation that neighbouring pixels with similar color values usually share the same semantic class, we introduce a Laplacian loss to ensure the consistency of the class predictions with the image contents. This prior knowledge is incorporated into the training of the GCN in the form of the Laplacian loss:
\begin{equation}
\label{eq:loss_lp}
 \ell_{lp} = \frac{1}{2|\mathcal{V}|} \sum_{i \in \mathcal{V}}\sum_{j \in \mathcal{V}}{\Phi_{i,j}\|q_i-q_j\|_2^2}, 
\end{equation}
 which aims to minimize the discrepancy (measured in $l_2$ norm) between the class prediction of pixel $i$ and those of its surrounding pixels $j$ in a neighborhood $\mathcal{N}_i$ according to the affinity weight $\Phi_{i,j}$ that reflects the similarity between pixel $i$ and pixel $j$ in terms of their color values and locations as given by
\begin{equation}
\label{eq:loss_lp_w}
\Phi_{i,j} = 
    \begin{cases}
    \exp(-\dfrac{|| \bar I_i-  \bar I_j||^2}{2\sigma_1^2 }
    -\dfrac{||f_i- f_j||^2}{2\sigma_2^2 }) & \mbox{if } j \in \mathcal{N}_i   \\ 
    0  & \mbox{otherwise}
    \end{cases},
\end{equation}
where $\bar I \in \mathbb{R}^{3 \times H/S  \times W/S }$ is the bilinearly downscaled version of the input image $I$; $\bar{I}_i$ and $f_i=(x_i,y_i)$ refer to the color value \textcolor{black}{at pixel $(x_i,y_i)$ and its $x,y$ coordinates}, respectively; \textcolor{black}{$\sigma_1=\sqrt{3}$, $\sigma_2=10$ are hyper-parameters}; and \textcolor{black}{$\mathcal{N}_i$ is a $5 \times 5$ window centered at pixel $(x_i,y_i)$}. Note that $\Phi$, which relies on low-level color and spatial information to regularize the GCN output, is to be distinguished from the affinity matrix $A$ in Eq.~(\ref{eq:GCN}), which uses high-level semantic information~\cite{PSA18,IRNet19} to specify the graph structure of GCN for feature propagation as will be detailed next. 

%% file: experiment.tex
\section{Experimental Results}   \label{sec:exp}
\subsection{Setup} \label{sec:setup}
\noindent \textbf{\textcolor{black}{Datasets and Metrics:}} \textcolor{black}{Following most of the prior works~\cite{PSA18,IRNet19,SEAM20,SCE20,SingleStage20,RRM20,MCIS20,WSSSGNN21}} we evaluate our method on the PASCAL VOC 2012 semantic segmentation benchmark~\cite{VOC10}. It includes images with pixel-wise class labels, of which 1,464, 1,449, and 1,456 are used for training, validation, and test, respectively. Following the common training strategy for semantic segmentation, we adopt augmented training images \cite{AUG}, which are composed of 10,582 images annotated by pixel-wise class labels. However, we use only image-level class labels for weak supervision during training. To evaluate the semantic segmentation accuracy, we resort to Mean Intersection over Union (mIoU) as the other baselines.

\noindent \textbf{\textcolor{black}{Implementation and Training:}}
There are two variants, termed WSGCN-I and WSGCN-P, of our model that use affinity matrices $A$, node features $\mathcal{V}$ and CAMs from IRNet~\cite{IRNet19} and PSA~\cite{PSA18}, respectively. The WSGCN-I constructs the affinity matrix $A$ using the boundary detection network~\cite{IRNet19} and takes as node features $\mathcal{V}$ the features that sit in the last layer of the boundary detection network and before its $1 \times 1$ convolutional layer. In contrast, the WSGCN-P follows AffinityNet~\cite{PSA18} in specifying the affinity matrix $A$ and uses the semantic features for affinity evaluation as node features $\mathcal{V}$. Both variants apply the same CAMs as their counterparts in IRNet~\cite{IRNet19} and PSA~\cite{PSA18}. 
For every training image, we train both variants for $250$ gradient update steps with a dropout rate of $0.3$. The $\beta_1,\beta_2$ in Eq.~(\ref{eq:loss_all}) are set \textcolor{black}{to $10$ and $10^{-2}$, respectively}. 
The learning rate of the optimizer Adaptive Moment Estimation (Adam) is set to $0.01$, with a weight decay of $5\times 10^{-4}$. For a fair comparison, we follow \cite{SCE20,SingleStage20,RRM20} to adopt DeepLabv2~\cite{Deeplab18} as our semantic segmentation network in the second stage. The backbone ResNet-101 is pre-trained on ImageNet.

\noindent \textbf{\textcolor{black}{Baselines:}} We compare our approach with several state-of-the-art weakly-supervised image semantic segmentation methods~\cite{PSA18,IRNet19,SEAM20,SCE20,SingleStage20,RRM20,MCIS20,WSSSGNN21} on training, validation, and test sets.
The comparison with PSA~\cite{PSA18} and IRNet~\cite{IRNet19} singles out the benefits of our feature propagation framework over label propagation since \textcolor{black}{our GCN adopts the same affinity matrix as theirs for propagating features rather than activation scores of CAMs}.  
\input{table/CPL_iou}
\subsection{Quality of Complete Pseudo Labels} \label{exp:ablation}
Table~\ref{tab:cpl} presents a quality comparison of complete pseudo labels between our WSGCN-I/P and several state-of-the-art methods \cite{PSA18,IRNet19,SEAM20,SCE20,SingleStage20,MCIS20,WSSSGNN21} in terms of mIoU on the PASCAL VOC 2012 training set. \textcolor{black}{Most of these baselines propagate activation scores of CAMs with random walk (i.e., label propagation) to infer complete pseudo labels}. 

We see that our WSGCN-I achieves the highest mIoU (68.0\%) among all the competing methods. It shows a 1.5\% mIoU improvement over IRNet~\cite{IRNet19}. Likewise, WSGCN-P has a 4.3\% mIoU gain as compared to PSA~\cite{PSA18}. These results confirm the superiority of our feature propagation framework over the label propagation adopted by these baselines. 

Further insights can be gained by noting that in generating complete pseudo labels for every training image, our schemes rely on training a specific GCN by back-propagation. On the contrary, label propagation is a feed-forward process. During back-propagation, our schemes involve not only high-level node features $\mathcal{V}$ in constructing an affinity matrix $A$ for feature propagation but also low-level color/spatial information (cp. $\Phi$ in $\ell_{lp}$) in regularizing label predictions, whereas the label propagation used by the baselines depends largely on high-level node features $\mathcal{V}$ to compute a Markov transition matrix. A side experiment shows that \textcolor{black}{substituting the affinity matrix $A$ in Eq.~(\ref{eq:GCN}) for $\Phi$ in $\ell_{lp}$} causes the mIoU of complete pseudo labels to decrease from $68\%$ to $66.3\%$. This is in agreement with the general observation that high-level semantic features usually contain less information about spatial details. Note also that the performance gap between WSGCN-I and WSGCN-P comes from the use of different affinity matrices $A$ and node features $\mathcal{V}$, emphasizing the influence of these design choices on the quality of complete pseudo labels. 
\subsection{Ablation Study of Loss Functions}
Table~\ref{tab:loss_bk} further analyzes how the single use of various loss functions and their combinations along with dCRF contribute to the quality of complete pseudo labels. By default, the foreground and background losses are enabled. They alone (WSGCN-I without any check mark) offer $62.0\%$ mIoU. The entropy loss improves the mIoU further by $1.5\%$. The Laplacian loss attains even higher gain ($2.4\%$) due to the incorporation of low-level color/spatial regularization. When combined, they show a synergy effect of $4.7\%$. 
 Recall that the entropy loss encourages the GCN to produce a one-hot-like semantic prediction while the Laplacian loss requires the predictions to be close to each other for adjacent pixels with similar colors. 
It is interesting to note that dCRF (as a post-processing step) can further improve on the entropy loss, the Laplacian loss as well as their combination. Although dCRF uses similar low-level information to the Laplacian loss, it functions as a separate, refinement step rather than a substitute for the Laplacian loss. Unlike dCRF, the Laplacian loss is included in the training loop of GCN. 

Fig.~\ref{fig:ent_lap} visualizes how these loss functions improve incrementally the quality of complete pseudo labels. We see that based on the semantic segmentation resulting from the foreground and background losses, the entropy loss tends to grow the foreground regions while the Laplacian loss can help alleviate semantic segmentation errors at object boundaries.           
\input{table/breakdown}
\input{table/exp_regular_tran}


\input{table/SOTA_all}
\input{table/Label_Pro}
\input{table/IOU_test}

\subsection{Semantic Segmentation Performance}
Using complete pseudo labels as supervision, we train a semantic segmentation network in the second stage. Table~\ref{tab:sota_all} compares the semantic segmentation accuracy of our 2-stage training framework with GCN-based feature propagation to several state-of-the-art methods.  We see that our WSGCN-I outperforms all the baselines \textcolor{black}{which do not use saliency maps} on both validation and test datasets, with the ResNet-101 backbone pre-trained on ImageNet. It gains 0.5\% (respectively, 2\%) more mIoU on the test (respectively, validation) dataset, when pre-training the backbone on MS-COCO dataset as in~\cite{SCE20, RRM20}. 


Table~\ref{tab:cmp_psa_ir} further compares WSGCN-P and WSGCN-I with their counterparts PSA~\cite{PSA18} and IRNet~\cite{IRNet19}, respectively. Note that the main difference between WSGCN-P and WSGCN-I is the choice of the affinity matrix (Section \ref{sec:setup}). For a fair comparison, we choose DRN-D-105~\cite{DRN17}, which is pre-trained on ImageNet and is publicly accessible for reproducibility, as the backbone in DeepLabv2. It is seen that these variants show $0.9\%$ to $2.5\%$ mIoU improvements, confirming the superiority of our feature propagation framework to label propagation. The results also demonstrate that our scheme can work with different affinity matrices.  

Table~\ref{tab:acc_test} additionally compares our WSGCN-I and IRNet~\cite{IRNet19} with the ResNet-101 backbone. As shown, WSGCN-I achieves higher IoU in most categories than IRNet~\cite{IRNet19} and has a $2.3\%$ mIoU improvement.

%% file: table/CPL_iou.tex
\newcommand{\tabincell}[2]{\begin{tabular}{@{}#1@{}}#2\end{tabular}}
\begin{table}
    \centering
    \caption{Quality comparison of complete pseudo labels on PASCAL VOC 2012 training set.}
    \setlength\tabcolsep{0.12cm}
    \begin{tabular}{l|c}
        \hline

        Method & mIoU (\%) \\ 
        \hline
        PSA~\cite{PSA18}   &59.7\\
        SC-CAM~\cite{SCE20} &63.4\\
        SEAM~\cite{SEAM20} &63.6\\
        IRNet~\cite{IRNet19} &66.5\\
        SingleStage~\cite{SingleStage20} &66.9\\
        \hline
        WSGCN-P & 64.0\\ 
        WSGCN-I & \textbf{68.0}\\
        \hline
        \end{tabular}
        \vspace{-0.8em}
    \label{tab:cpl}

\end{table}

%% file: table/breakdown.tex
\begin{table}
    \centering
    \caption{Effect of various loss functions on the quality of complete pseudo labels evaluated on PASCAL VOC 2012 training set.}
    \setlength\tabcolsep{0.12cm}
    \begin{tabular}{l|ccc|c}
        \hline
        \multirow{ 2}{*}{Method}  & Entropy & Laplacian & dCRF & \multirow{ 2}{*}{mIoU (\%)} \\ 
        & loss & loss &  (post-processing) &  \\
        \hline
        WSGCN-I & &  &                 & 62.0\\
        WSGCN-I &$\surd$&&              & 63.5\\
        WSGCN-I &       &$\surd$&       & 64.4\\
        WSGCN-I &$\surd$&$\surd$&       & 66.7\\
        \hline
        WSGCN-I &$\surd$&       &$\surd$& 66.0\\
        WSGCN-I &       &$\surd$&$\surd$& 66.1\\
        WSGCN-I &$\surd$&$\surd$&$\surd$& \textbf{68.0}\\
        \hline
        \end{tabular}
    \label{tab:loss_bk}

\end{table}

%% file: table/exp_regular_tran.tex
\begin{figure}[t]
    \small
    \centering
    \setlength\tabcolsep{0.01cm}
    \renewcommand{\arraystretch}{0.5}
   
    \begin{tabular}{rcccccc}
        \rotatebox{90}{\parbox{2cm}{\begin{center}Partial pseudo label\end{center}}}&
        \includegraphics[width=1.8cm,height=1.801cm]{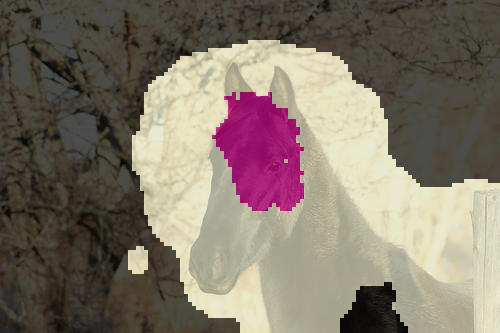}&
        \includegraphics[width=1.8cm,height=1.801cm]{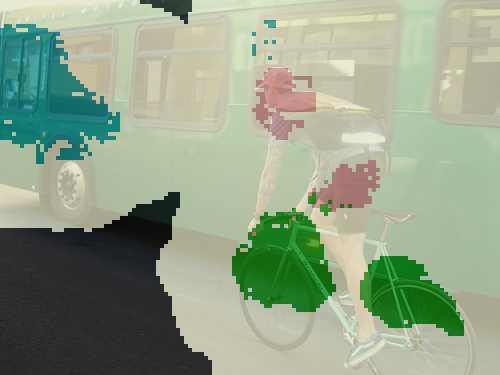}&
        \includegraphics[width=1.8cm,height=1.801cm]{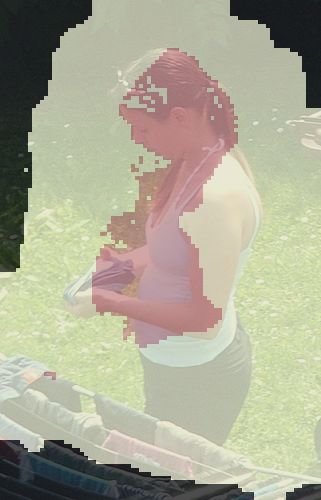}&
        \includegraphics[width=1.8cm,height=1.801cm]{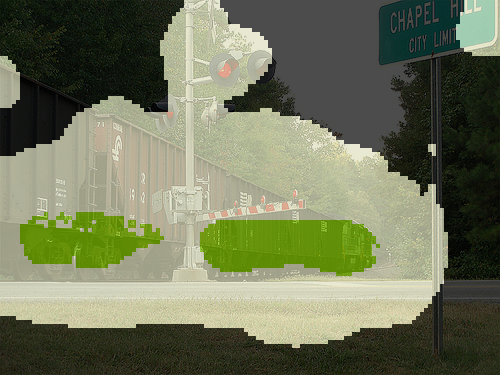}
        \\
        \rotatebox{90}{\parbox{1.801cm}{\begin{center}$\ell_{fg}+\ell_{bg}$\end{center}}} &
        \includegraphics[width=1.8cm,height=1.801cm]{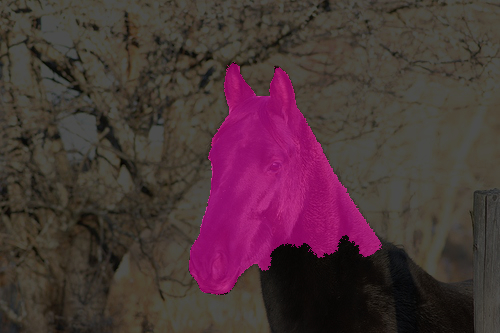} &
        \includegraphics[width=1.8cm,height=1.801cm]{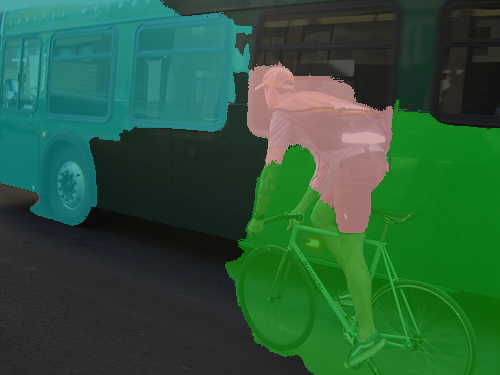} &
        \includegraphics[width=1.8cm,height=1.801cm]{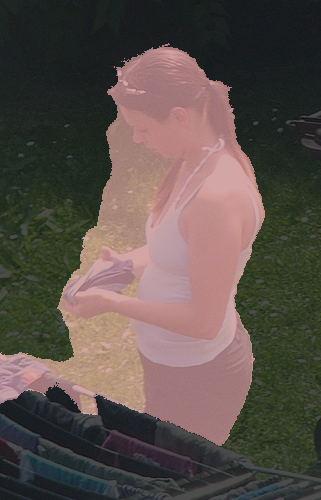}&
        \includegraphics[width=1.8cm,height=1.801cm]{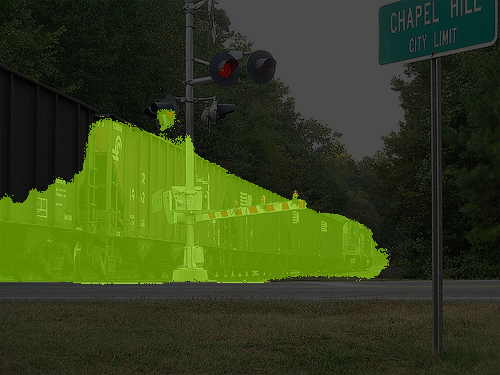}
        \\
        \rotatebox{90}{\parbox{1.801cm}{\begin{center}$\ell_{fg/\/bg}+\ell_{ent}$\end{center}}} &
        \includegraphics[width=1.8cm,height=1.801cm]{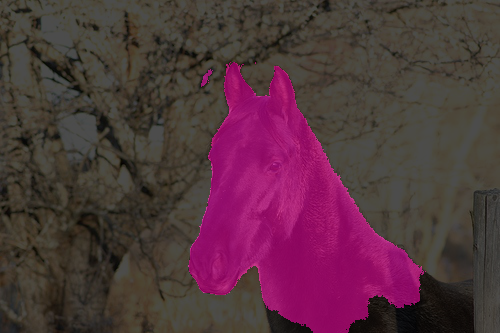} &
        \includegraphics[width=1.8cm,height=1.801cm]{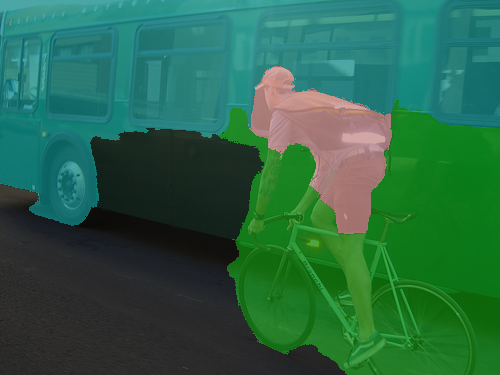} &
        \includegraphics[width=1.8cm,height=1.801cm]{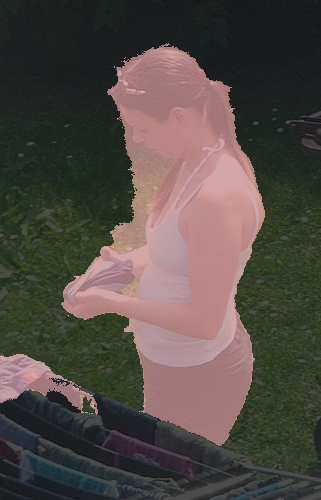}&
        \includegraphics[width=1.8cm,height=1.801cm]{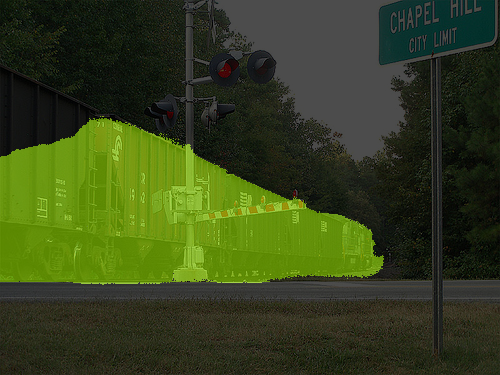}
        \\
        \rotatebox{90}{\parbox{1.801cm}{\begin{center}WSGCN-I\end{center}}} &
        \includegraphics[width=1.8cm,height=1.801cm]{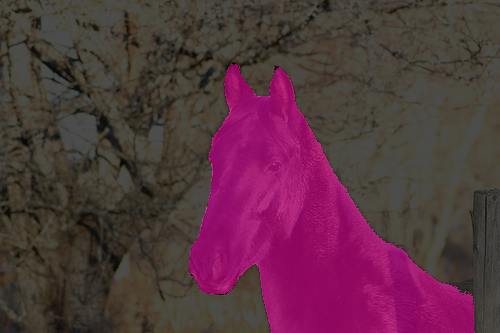}&
        \includegraphics[width=1.8cm,height=1.801cm]{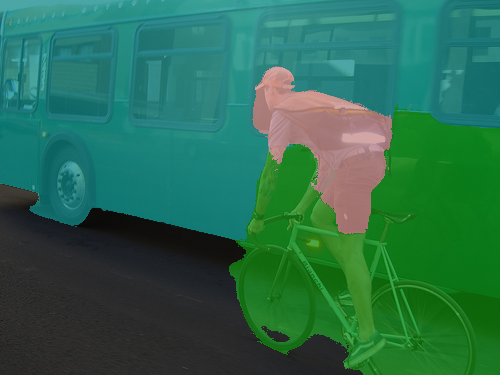}&
        \includegraphics[width=1.8cm,height=1.801cm]{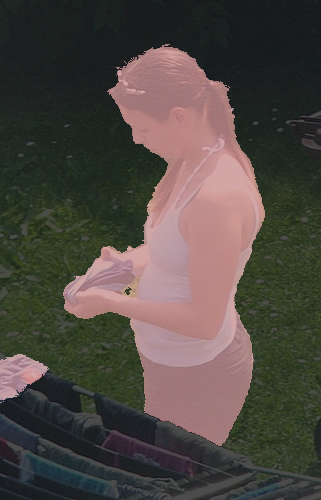}&
        \includegraphics[width=1.8cm,height=1.801cm]{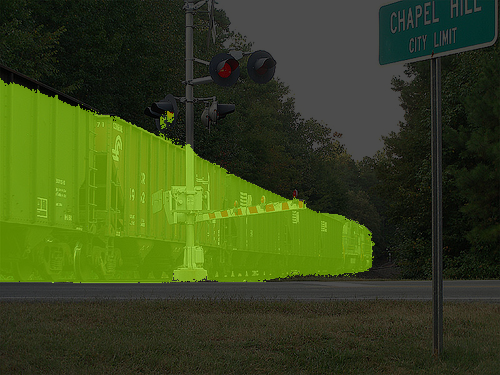}
        \\
        \rotatebox{90}{\parbox{1.801cm}{\begin{center}IRNet\end{center}}}&
        \includegraphics[width=1.8cm,height=1.801cm]{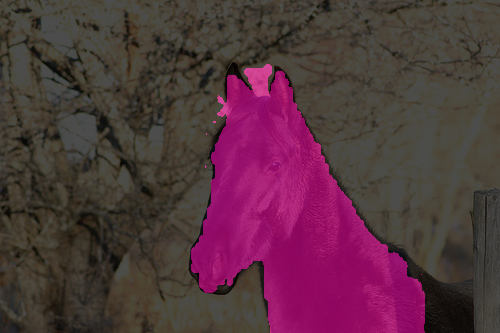} &
        \includegraphics[width=1.8cm,height=1.801cm]{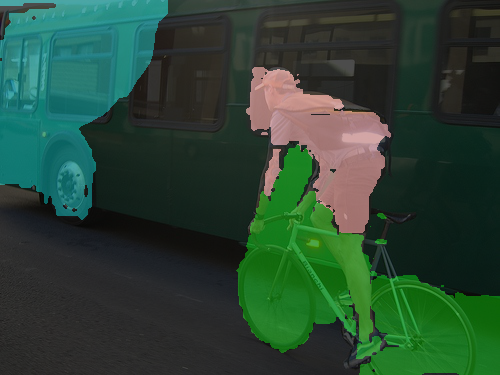} &
        \includegraphics[width=1.8cm,height=1.801cm]{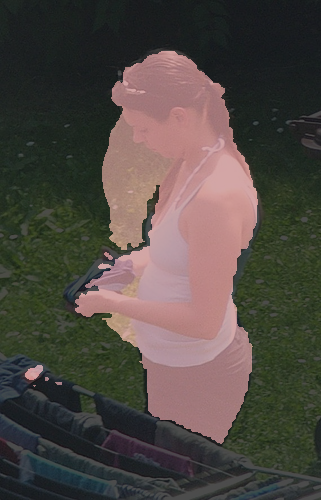} &
        \includegraphics[width=1.8cm,height=1.801cm]{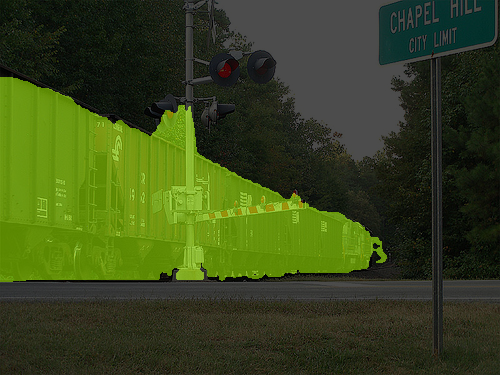}
        \\
        \rotatebox{90}{\parbox{1.801cm}{\begin{center}GT\end{center}}}&
        \includegraphics[width=1.8cm,height=1.801cm]{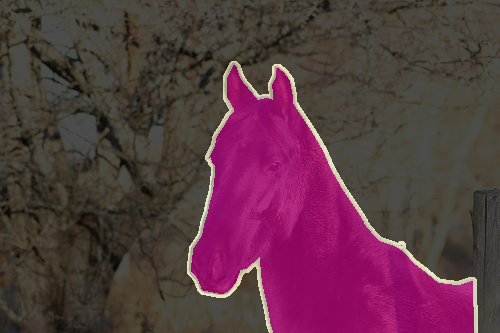} &
        \includegraphics[width=1.8cm,height=1.801cm]{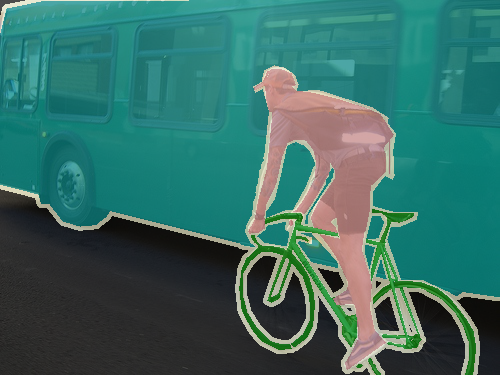} &
        \includegraphics[width=1.8cm,height=1.801cm]{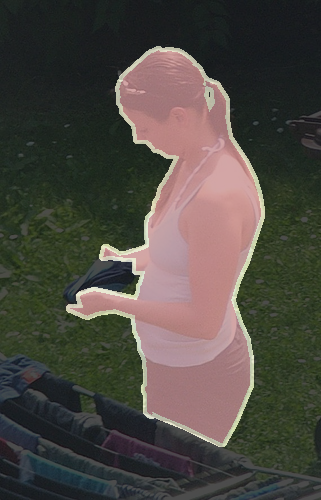} &
        \includegraphics[width=1.8cm,height=1.801cm]{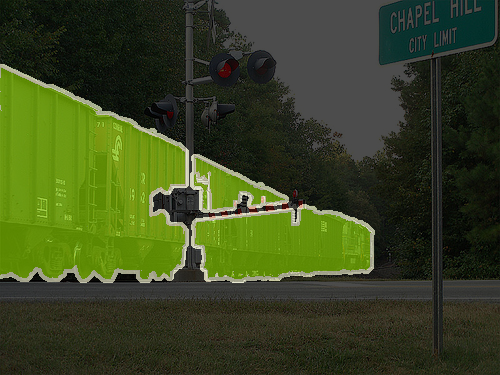}
        \\
    \end{tabular}
    
    \caption{Sample results of pseudo labels. From top to bottom: partial pseudo labels where white regions indicate unlabeled pixels, complete pseudo labels by $\ell_{fg}+\ell_{bg}$, complete pseudo labels by $\ell_{fg}+\ell_{bg}+\ell_{ent}$, complete pseudo labels by all four losses (WSGCN-I), complete pseudo labels by IRNet, and the ground-truths. All complete pseudo labels are refined with dCRF.}
    \vspace{-0.8em}
    \label{fig:ent_lap}
\end{figure}

%% file: table/SOTA_all.tex
\begin{table}
    \centering
    \caption{Comparison of semantic segmentation accuracy evaluated on PASCAL VOC 2012 dataset. 
    }
    \setlength\tabcolsep{0.1cm}
    \begin{tabular}{lcc|cc}
        \hline
        \multirow{ 2}{*}{Method}&\multirow{ 2}{*}{Publication} &\multirow{ 2}{*}{Backbone} & \multicolumn{2}{c}{mIoU (\%)} \\
         &  &  & $val$ & $test$ \\
        
        \hline
        
        PSA~\cite{PSA18}            &CVPR18&  ResNet-38 &  61.7  & 63.7 \\
        IRNet~\cite{IRNet19}        &CVPR19&  ResNet-50 &  63.5  & 64.8 \\
        SEAM~\cite{SEAM20}          &CVPR20&  ResNet-38 & 64.5 & 65.7 \\
        \hline
        SC-CAM$\dagger$~\cite{SCE20} &CVPR20&  ResNet-101 & 66.1 & 65.9\\
        RRM$\dagger$~\cite{RRM20}      &AAAI20&  ResNet-101 & 66.3 & 66.5\\
        SingleStage$\dagger \dagger$~\cite{SingleStage20}    &CVPR20&  ResNet-101& 65.7& 66.6\\
        MCIS*~\cite{MCIS20}       &ECCV20& ResNet-101 &  66.2& 66.9\\
        GroupWSSS*~\cite{WSSSGNN21}       &AAAI21& ResNet-101 &  68.2& 68.5\\
        
        \hline
        \textbf{WSGCN-I (ours)}       &--&  ResNet-101 &  66.7 & 68.8 \\
        \textbf{WSGCN-I$\dagger$ (ours)}       &--&  ResNet-101 &  \textbf{68.7} & \textbf{69.3} \\
        \hline
        \multicolumn{5}{l}{*:~Using saliency maps as extra supervision signals.}\\ 
        \multicolumn{5}{l}{$\dagger$:~Pre-training the backbone in DeepLabv2 on MS-COCO.}\\
        \multicolumn{5}{l}{$\dagger \dagger$:~Using DeepLabv3+ as the semantic segmentation model.}\\
    \end{tabular}
    \vspace{-0.8em}
    \label{tab:sota_all}
\end{table}

%% file: table/Label_Pro.tex
\begin{table}[t!]
    \centering
    \caption{Comparison of PSA, IRNet, and our method with \textcolor{black}{DRN-D-105 backbone}. 
    }
    \setlength\tabcolsep{0.2cm}
    \begin{tabular}{lc|cc}
        \hline
        \multirow{ 2}{*}{Method} & \multirow{ 2}{*}{Backbone} & \multicolumn{2}{c}{mIoU (\%)} \\
         &  & $val$ & $test$ \\
        \hline
         PSA~\cite{PSA18}   &DRN-D-105 & 60.9  & 61.9 \\
         \textbf{WSGCN-P (ours)} & DRN-D-105 & \textbf{63.1} & \textbf{64.4}   \\
         \hline
         IRNet~\cite{IRNet19}  &DRN-D-105 & 66.0  & 66.5 \\
         \textbf{WSGCN-I (ours)} & DRN-D-105 &\textbf{66.9}& \textbf{67.5}   \\
        \hline
        
    \end{tabular}
    \vspace{-0.8em}
    \label{tab:cmp_psa_ir}
\end{table}

%% file: table/IOU_test.tex
\begin{table*}
    \footnotesize
    \caption{\textcolor{black}{Comparison of WSGCN-I and IRNet with the ResNet-101 backbone on the PASCAL VOC 2012 test set.}}
    \setlength\tabcolsep{3.1pt}
    \begin{tabular}{l|ccccccccccccccccccccc|c}
        \hline
        Method & bkg & aero & bike & bird & boat & bot. & bus & car & cat & chair & cow & tab. & dog & horse & mbk & per. & plant & sheep & sofa & train & tv &  mIoU \\
        \hline
        IRNet~\cite{IRNet19} &90.2 &75.8 &32.5 &76.5 &49.6 &\textbf{65.9} &82.6 &76.2 &82.7 & \textbf{31.8} & \textbf{77.5} &41.8 &79.4 &75.0 &81.2 & \textbf{75.2} &47.9 &81.1 & \textbf{51.5} &62.5 & \textbf{59.1} &66.5\\
        \hline
        \textbf{WSGCN-I} & \textbf{91.0} & \textbf{79.8} & \textbf{33.8} & \textbf{78.2} & \textbf{50.9} & 65.7 & \textbf{86.8} & \textbf{79.3} & \textbf{86.0} & 27.4 & 75.2 & \textbf{48.9} & \textbf{83.4} & \textbf{76.2} & \textbf{82.7} & 74.4 & \textbf{64.3}  & \textbf{86.0}  & 51.0  & \textbf{64.2} & 58.6  & \textbf{68.8}   \\
        \hline
    \end{tabular}
    \label{tab:acc_test}
\end{table*}


%% file: conclusion.tex
\section{Conclusion}
\label{section:conclusion}
This paper presents a GCN-based feature propagation framework for weakly-supervised image semantic segmentation. Unlike feed-forward label propagation, our approach optimizes the generation of complete pseudo labels by learning off-line a separate GCN for every training image. The learning process involves feature propagation based on high-level semantic affinity between pixels and label prediction regularized by low-level spatial coherence. On our computation platform with 4 NVIDIA \textcolor{black}{Tesla V100} GPUs and 1 Intel Xeon 3GHz CPU, \textcolor{black}{the runtime needed to infer complete pseudo labels for one sample image is about 2.45s, as compared to 1.43s for label propagation with IRNet~\cite{IRNet19}.} Experimental results on PASCAL VOC 2012 benchmark validate the superiority of our framework to label propagation.
